\documentclass[journal]{IEEEtran}
\ifCLASSINFOpdf
  \usepackage[pdftex]{graphicx}
  \graphicspath{{../pdf/}{../jpeg/}}
\else
  \usepackage[dvips]{graphicx}
  \graphicspath{{../eps/}}
\fi

\ifCLASSOPTIONcompsoc
 \usepackage[caption=false,font=normalsize,labelfont=sf,textfont=sf]{subfig}
\else
 \usepackage[caption=false,font=footnotesize]{subfig}
\fi

\ifCLASSINFOpdf
  \usepackage[pdftex]{graphicx}
\else
  \usepackage[dvips]{graphicx}
\fi

\usepackage{amsmath}
\usepackage{tikz}
\usetikzlibrary{shapes,arrows}

\usepackage{algpseudocode}
\usepackage{multirow}

\makeatletter
\renewcommand{\ALG@beginalgorithmic}{\footnotesize}
\makeatother

\begin{document}
\title{Rotation-invariant shipwreck recognition \\ with forward-looking sonar}

\author{Gustavo~Neves, R\^omulo~Cerqueira, Jan~Albiez and~Luciano~Oliveira
\thanks{
    G. Neves and R. Cerqueira are with the Brazilian Institute of Robotics, at SENAI-CIMATEC, Bahia, Brazil (see http://www.senaicimatec.com.br/en). FlatFish project is part of an ongoing research of BG Group Brasil, SENAI-CIMATEC, Ag\^{e}ncia Nacional do Petr\'{o}leo, G\'{a}s Natural e Biocombust\'{i}veis (ANP) and EMPRAPII (Empresa Brasileira de Pesquisa e Inova\c{c}\~{a}o Industrial) program, under the grant.
}
\thanks{
    J. Albiez is with the German Research Center for Artificial Intelligence (DFKI), Bremen, Germany (https://www.dfki.de).
}
\thanks{
    L. Oliveira is with Intelligent Vision Research Lab, at Federal University of Bahia (UFBA), Bahia, Brazil (http://www.ivisionlab.dcc.ufba.br).
}
}


\markboth{}%
{Neves \MakeLowercase{\textit{et al.}}: Rotation-invariant shipwreck detection}

\maketitle

\begin{abstract}
Under the sea, visible spectrum cameras have limited sensing capacity, being able to detect objects only in clear water, but in a constrained range. Considering any sea water condition, sonars are more suitable to support autonomous underwater vehicles' navigation, even in turbid condition. Despite that sonar suitability, this type of sensor does not provide high-density information, such as optical sensors, making the process of object recognition to be more complex. To deal with that problem, we propose a novel trainable method to detect and recognize (identify) specific target objects under the sea with a forward-looking sonar. Our method has a preprocessing step in charge of strongly reducing the sensor noise and seabed background. To represent the object, our proposed method uses histogram of orientation gradient (HOG) as feature extractor. HOG ultimately feed a multi-scale oriented detector combined with a support vector machine to recognize specific trained objects in a rotation-invariant way. Performance assessment demonstrated promising results, favoring the method to be applied in underwater remote sensing.
\end{abstract}


\IEEEpeerreviewmaketitle

\section{Introduction}

\IEEEPARstart{U}{nderwater} facilities of oil and gas fields must be periodically inspected with the goal of investigating the condition of submerged structures. The very first goal with this task is to verify the need of repair and maintenance, being these tasks performed by remotely operated vehicles (ROV) or divers. These inspections are complex, expensive and manually carried out, demanding a complete support, usually comprised of a crane, umbilical cable and the ROV crew. 

Because of this expensive cost of oil and gas facility maintenance, many researches have been developing autonomous underwater vehicles (AUVs) to be applied in the aforementioned tasks. AUVs aim at conducting survey missions, using internal and external sensing devices with lower operational costs. The vehicle returns to a pre-programmed location when a mission is completed, and the gathered data can be downloaded and analyzed. One of the underlying AUV's tasks is to detect submerged objects, providing reference locations to support the vehicle navigation. 

The use of visible spectrum cameras in underwater inspection-driven AUVs is limited by turbid or deep waters, while imaging sonars have been exploited to coverage large areas even under low-to-zero visibility conditions. Sonars emit sound waves in a given direction until these waves hit with an object, having part of the wave energy reflected back. By calculating the time-of-flight of the sound waves, a distance between the sonar and the target is established. On measuring the backscattered energy, it is possible to define the target shape, and each record of sliced time is called bin. All bins scanning in the same direction angle composes a beam. While sonars are more independent with respect to water turbidity conditions, these sensors usually provide a more difficult data interpretation because of the acquisition and environment characteristics.

Some works on sonar image to support AUV's underwater navigation have been developed. Usually these works attempt to eliminate the seabed, tracking highlighted areas without the ability of recognizing the target object. Cuschier and Negahdaripous \cite{Cuschieri1998} adapted the optical flow method to track any feature from a forward-scan sonar images; they used the sonar image intensities to estimate the motion parameters; the goal is to help the AUV navigate underwater. Ruiz \textit{et al.} \cite{TenaRuiz1999} perform image processing techniques to detect submerged objects, tracking them with a Kalman filter; they segment objects in multi-beam sonar images using a region growing algorithm; the \textit{position}, the \textit{orientation} and the \textit{area} from the segmented areas are used as features to track the target objects. Petillot \textit{et al.} \cite{Petillot2001} segment underwater objects, and extract features for AUVs' obstacle avoidance and path planning; image are segmented by an adaptive threshold technique, and posteriorly features, such as area, perimeter and moments, are extracted from the segmented regions, and a Kalman filter is used to track the obstacles. Folkesson \textit{et al.} \cite{Folkesson2007} segment the sonar image, and the centroid positions of the segmented blobs is used to track the objects by using a probability hypothesis density (PHD) filter. Johansson \textit{et al.} \cite{Johannsson2010} extract dense features from a forward-looking sonar, applying pair-wise registration between consecutive sonar frames; sonar image registrations are combined with sensor information to improve vehicle navigation; features are extracted with the image gradient followed by an adaptive thresholding. Weng \textit{et al.} \cite{Weng2012} modified the Otsu threshold in order to separate the background and the foreground from the multi-beam sonar images; they built a color and area models in the first sonar frame, which is used to track the objects by particle filter based on multi-feature adaptive fusion. Yan \textit{et al.} \cite{yan2014} uses a different approach to detect object avoiding energy emission; by means of gradiometer, the gravity gradient differential ratio is measured, and used to estimate the objects body-center location and mass. Once these aforementioned methods do not distinguish one object from another, they do not allow performing accurate inspection tasks for a specific submerged object; also, used AUVs can not exploit the location of the inspected object as a reference for the navigation system, in these methods.

Different from the other works, the goal of this letter is to introduce a novel trainable method to detect and recognize (identify) specific under-the-sea target objects with a forward-looking sonar equipped in an AUV. Our contributions are: (i) a background reducing method, emphasizing the object shape in the sonar images, (ii) a multi-scale orientation detector, and (iii) a rotated-invariant recognition. The first contribution is achieved by applying image processing techniques in sonar images in order to reduce the sensor noise and seabed background; the second and the third contributions are achieved by applying image pyramid and sliding windows combined with support vector machine (SVM) over several sonar image orientations in order to search for the target object in a rotated-invariant way.

The proposed method was developed to be used in the FlatFish robot \cite{Albiez2015}, which is an AUV designed to perform inspection tasks in underwater facilities of oil and gas industry. Our method will support the navigation system to guide the vehicle by using reference recognized objects. 

\begin{figure}[t]
\centering
\subfloat[]{\includegraphics[width=0.49\columnwidth, height=3cm]{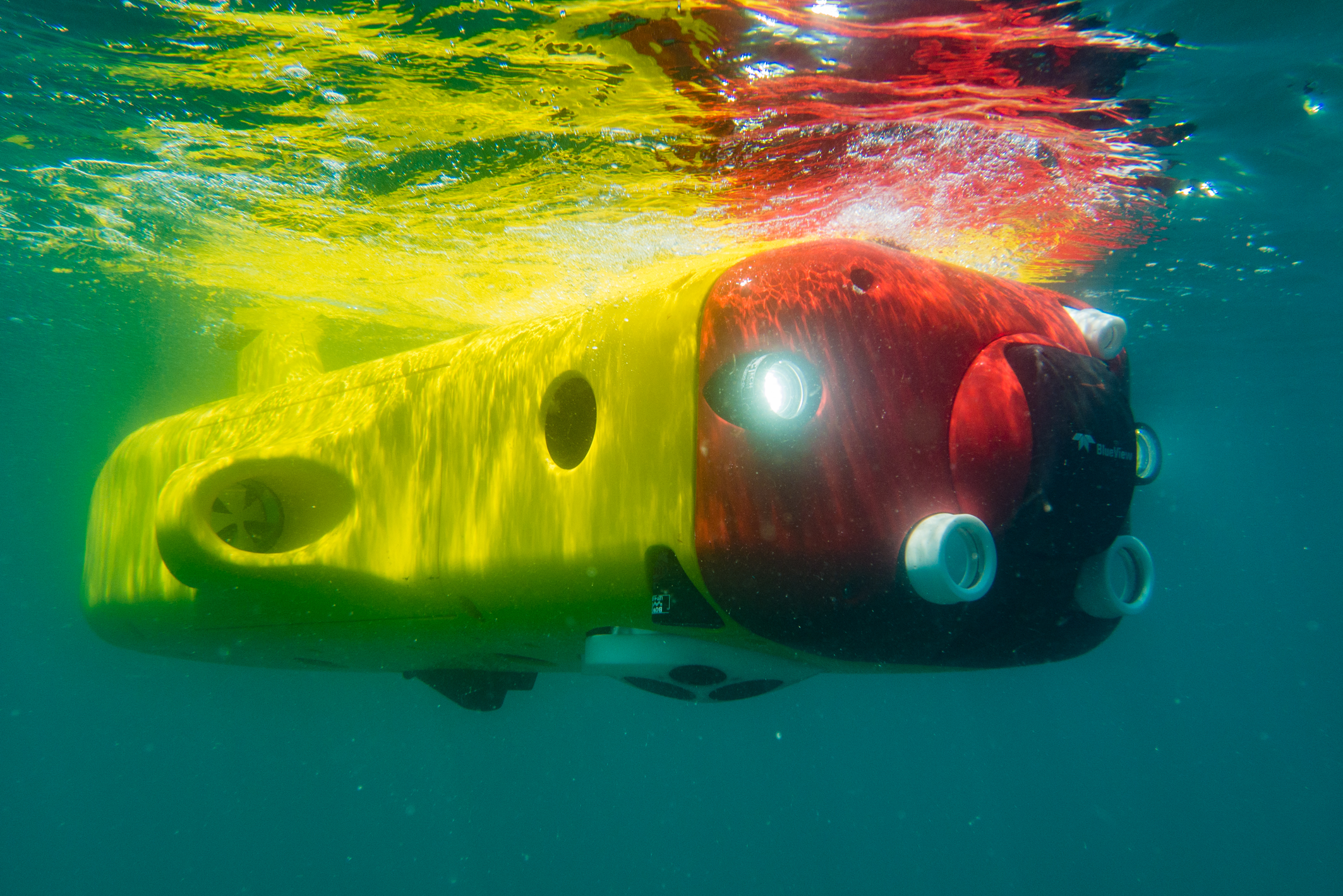}
\label{fig_flat_fish}}
\subfloat[]{\includegraphics[width=0.49\columnwidth, height=3cm]{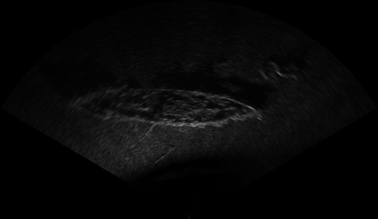}
\label{fig_raw_image}}
\captionsetup{justification=centering}
\caption{\protect\subref{fig_flat_fish} FlatFish robot and \protect\subref{fig_raw_image} \textit{Vapor da Jequitaia} acoustic visualization.}
\label{fig_flat_fish_and_shipwreck}
\end{figure}

\begin{figure}[t]
\centering
\includegraphics[width=0.3\textwidth]{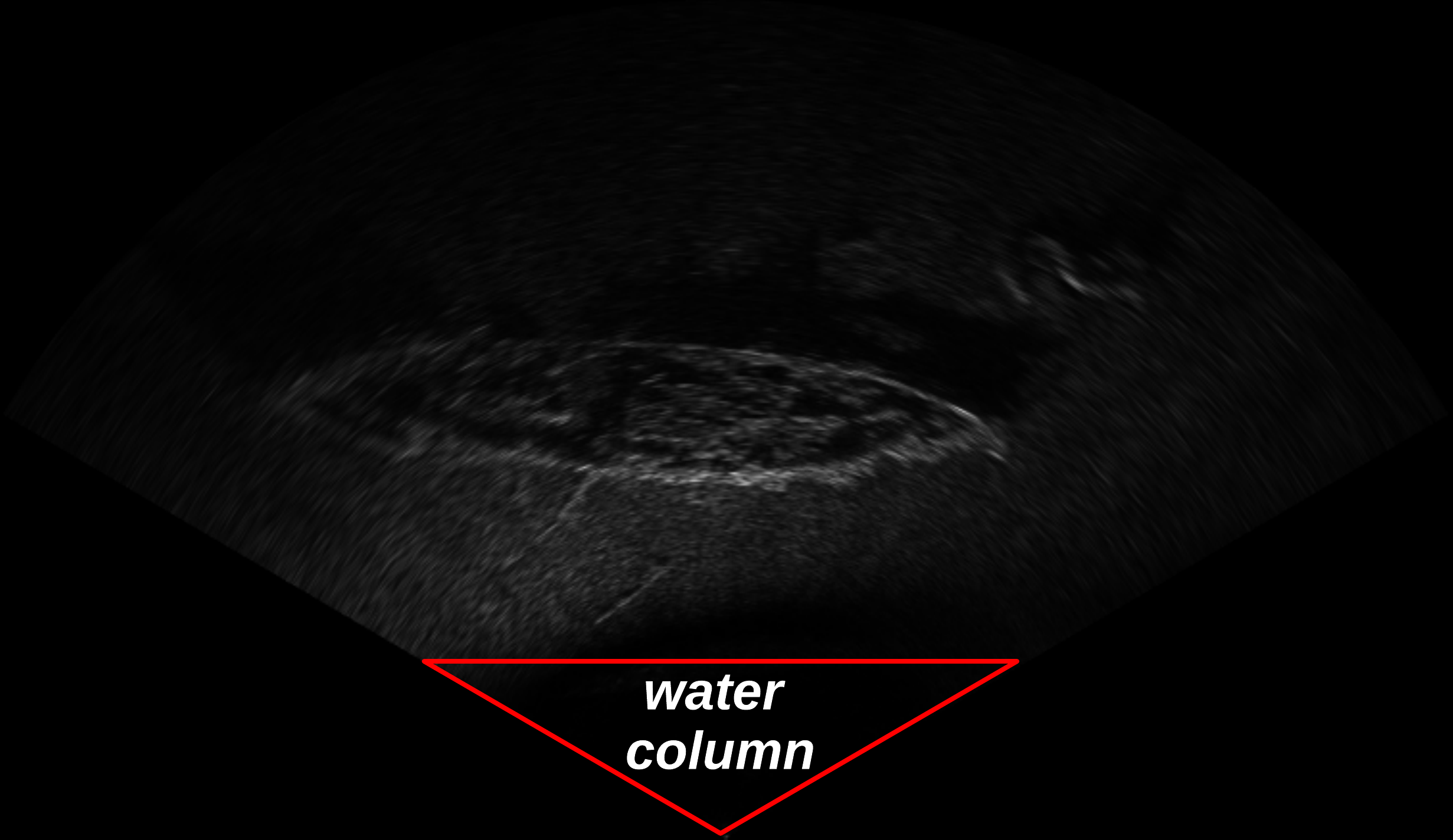}
\caption{Water column in the acoustic image.}
\label{fig_water_column}
\end{figure}

\begin{figure}[!t]
\centering
\subfloat[line average]{\includegraphics[width=0.49\columnwidth]{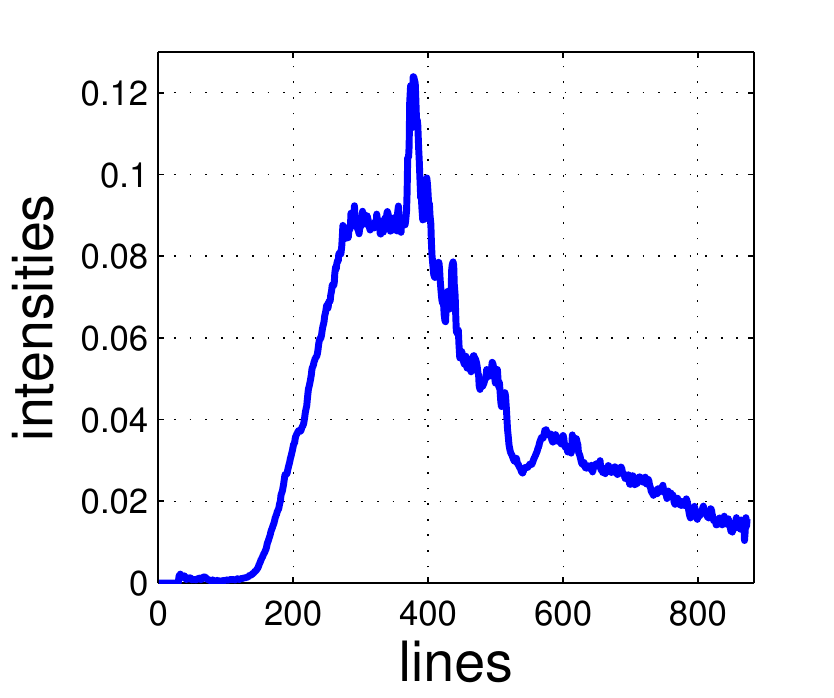}
\label{fig_line_means}}
\subfloat[cumulative sum]{\includegraphics[width=0.49\columnwidth]{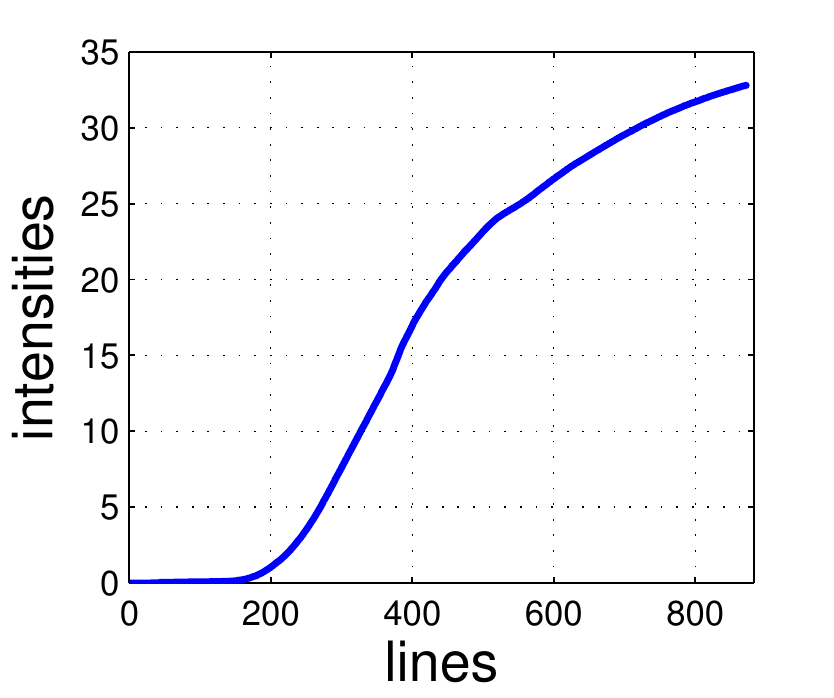}
\label{fig_accumulative_sum}}
\caption{Estimation of the starting point of the seabed. \protect\subref{fig_line_means} plot with the line averages. \protect\subref{fig_accumulative_sum} plot showing the cumulative sum of the line averages.}
\end{figure}

\section{Rotation-Invariant Shipwreck Detection}

\subsection{Preparing the sonar image}

\begin{figure*}[t]
\centering
\subfloat[]{\includegraphics[width=0.5\columnwidth, height=3cm]{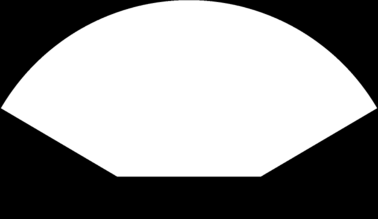}
\label{fig_region_of_interest}}
\subfloat[]{\includegraphics[width=0.5\columnwidth, height=3cm]{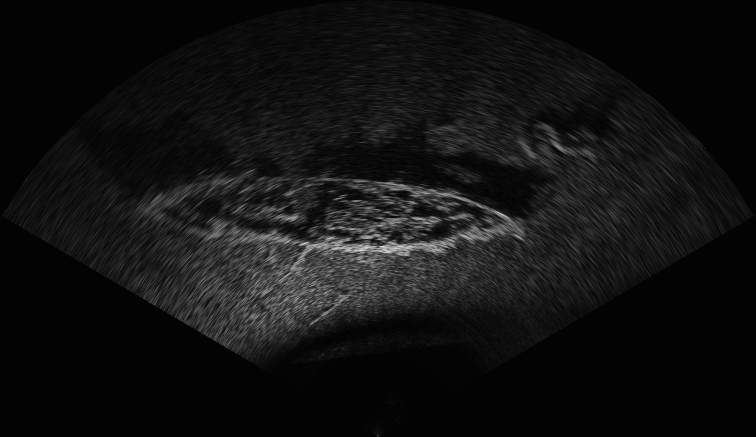}
\label{fig_enhanced}}
\subfloat[]{\includegraphics[width=0.5\columnwidth, height=3cm]{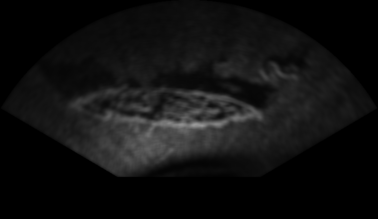}
\label{fig_noise_reduce}}
\subfloat[]{\includegraphics[width=0.5\columnwidth, height=3cm]{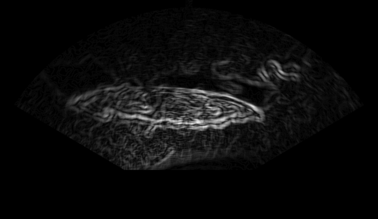}
\label{fig_edge_detection}}
\hfil
\subfloat[]{\includegraphics[width=0.5\columnwidth, height=3cm]{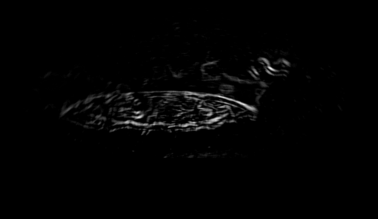}
\label{fig_mean_difference_filter}}
\subfloat[]{\includegraphics[width=0.5\columnwidth, height=3cm]{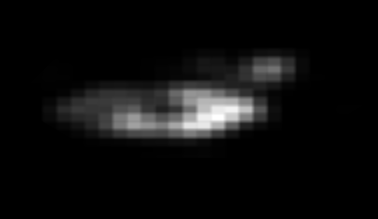}
\label{fig_saliency}}
\subfloat[]{\includegraphics[width=0.5\columnwidth, height=3cm]{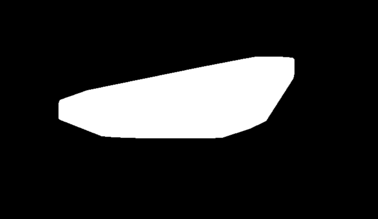}
\label{fig_preprocessing_mask}}
\subfloat[]{\includegraphics[width=0.5\columnwidth, height=3cm]{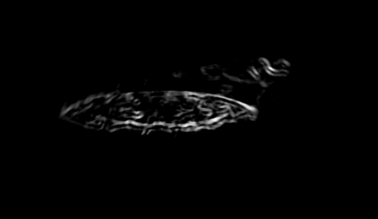}
\label{fig_preprocessing}}
\caption{Result of the preprocessing steps.
\protect\subref{fig_region_of_interest} the region of interest,
\protect\subref{fig_enhanced} image enhancement,
\protect\subref{fig_noise_reduce} noise reduction,
\protect\subref{fig_edge_detection} edge detection,
\protect\subref{fig_mean_difference_filter} difference of mean filters,
\protect\subref{fig_saliency} saliency map,
\protect\subref{fig_preprocessing_mask} preprocessed mask,
\protect\subref{fig_preprocessing} final preprocessed image,
}
\label{fig_preprocessing_filters}
\end{figure*}

The farther is the water column between the flatfish and the seabed, the bigger is the black region in the bottom of the sonar image (see Fig. \ref{fig_water_column}). That black region adversely affects the acoustic image processing, disturbing detection results. That region does not carry any information, so that it can be removed. This way, the first preprocessing step is to estimate the sonar range containing the beginning of the seabed, eliminating the black region created by the water column. To eliminate that, we first calculate the average of the image intensity over the sonar image lines (see Fig. \ref{fig_line_means}); next, the cumulative sum of the average of the image intensity is calculated to help determine the end of the water column (beginning of the seabed), which is represented by the point in the plot that the curve starts raising abruptly (close to 200, in Fig. \ref{fig_accumulative_sum}). After finding the end of the water column, an actual sonar region-of-interest is created without the black region, as depicted in Fig. \ref{fig_region_of_interest}.

Due to sonar configuration and the underwater terrain curvature, sonar image can present non-uniform illumination patterns. One such common pattern is the decrease of intensity values for bins far from the origin of the sonar pulse. This phenomenon occurs because the acoustic waves of the farthest bins travel greater distances than those of the closest bins. Hence the energy loss, caused by the transmission medium, is greater for the bins farthest from the origin of the pulse. To improve the illumination condition caused by that problem, we applied a modified version of a contrast limited adaptive histogram equalization (CLAHE) filter. We first calculate the image entropy in order to determine if the histogram is equalized, according to the following steps:

\begin{algorithmic}
\State $dst \gets src$
\State $C \gets C_{min}$
\State $H_0 \gets \textbf{entropy}(dst)$
\While{$C \geq C_{max}$}
\State $dst \gets \textbf{clahe}(dst, C, N)$
\State $H_1 \gets \textbf{entropy}(dst)$
\If{$ H_1 \geq H_{max}$ \textbf{ or } $H_0>H_1$ \textbf{ or } $|H_1-H_0|<H_{min\Delta}$}
\State \textbf{break}
\EndIf
\State $H_0 \gets H_1$
\State $C \gets C+C_{step}$
\EndWhile
\label{algo1}
\end{algorithmic}
where $src$ and $dst$ represent the source image and the result of the image enhancement, respectively. 

Toward finding the best clip limit, CLAHE is applied with a range of clip limits from $C_{min}$ to $C_{max}$, incremented by $C_{step}$. The current clip limit is represented by $C$ and the grid size is $N$. The entropy is calculated for each CLAHE result, and the best clip limit is the first one which satisfies any of the following conditions:

\begin{itemize}
\item $H_1 \geq H_{max}$, where $H_1$ is the current entropy and $H_{max}$ is the maximum entropy value;
\item $H_0>H_1$, where $H_0$ is the last entropy;
\item $|H_0-H_1|<H_{min\Delta}$, where $H_{min\Delta}$ is the minimum entropy difference.
\end{itemize}

The result of the image enhancement is shown in Fig. \ref{fig_enhanced}. After the image enhancement, the intensities far from the origin of the sonar are highlighted, enhancing the step of features extraction.

As it happens with other acoustic devices, the image acquired from the sonar suffers from low signal-noise rate (SNR). This primarily occurs due to the presence of speckle noise, which appears as a granular pattern in the acoustic images. Speckle noise is caused by the acoustic nature of the imaging sonar \cite{Cho2015}, adding high frequency components to the acoustic image, and decreasing the intensity of important information (such as shapes and object edges) \cite{Jaybhay2015}. Therefore the acoustic image is smoothed using a mean filter that aims at reducing the intensity of high frequency components. The neighborhood size used in our work is $11 \times 11$ pixels. An example of a resulting image in this step is shown in Fig. \ref{fig_noise_reduce}.

Objects found in the acoustic image are typically characterized by high intensities pixels, followed by shadows. This usually occurs due to the occlusion caused by the object itself (no echo is returned from occluded areas). Thus, object shapes in the sonar are seen as sharp transitions of intensities. In order to emphasize these sharp transitions, we apply an edge detection method, which uses vertical and horizontal Sobel derivatives to calculate the image gradient. Object shape is then emphasized, and the image background is reduced, as depicted in Fig. \ref{fig_edge_detection}.

Although the edge detection significantly decreases the image background, some edge components from seabed terrain and acoustic reverberation remain in the image. To tackle that problem, we applied the mean filter with two differently sized windows. Then the mean filter result from the larger window is subtracted from the result of the shorter window. To speed the computation of the means, the integral image is calculated. After the mean filters' subtraction, most of the pixels from background have negative values. These negative values are then set to zero, removing most of the pixels from the image background. We use a smaller window size of $3\times3$ pixels, and a larger window size of $50\times50$ pixels in our particular application. The result of this step is shown in Fig. \ref{fig_mean_difference_filter}.

To extract the regions containing objects, we developed a saliency map based on \cite{Achanta2008}. Our method divides the image into equally sized blocks. Next the average of each block is calculated using the integral image. The saliency map result for block $i$ is given by:

\begin{equation}
\small
D_{i}=\sum_{j=0}^{N} |\overline{R}_i-\overline{R}_j|
\label{eq_saliency_map_diffence}
\end{equation}
where $\overline{R}$ is the average of the block and $N$ is the total number of blocks. $D_i$ is the difference between block $i$ and the rest of the image. As the majority of the image contains background, blocks with highest difference values are considered as an object or part of it. The result is shown in Fig. \ref{fig_saliency}. The block size used was $24\times24$ pixels.

The saliency map result is segmented by the traditional Otsu method \cite{Otsu1979}. The convex hull contours is found by using the methods proposed by \cite{Sklansky1982} and \cite{Suzuki1985}, creating a final mask circumventing the objects (see Fig. \ref{fig_preprocessing_mask}). This mask represents the regions of the acoustic image where objects are expected to be. After the preprocessing, the shapes of objects are highlighted, the noise is decreased and the image background is reduced, as seen in Fig. \ref{fig_preprocessing}.

\subsection{Representing the object to be recognized}

To recognize the target object pattern, we applied histograms of oriented gradients (HOG) \cite{Dalal2004}, as feature
descriptors. The image result of the HOG descriptor extracted from the preprocessed image is seen in Fig. \ref{fig_hog_descriptor}, where lines are drawn to represent the strengths of the edge orientation in the histogram, for each cell.

\subsection{Learning the target object}
\label{sec_learning_process}

\begin{figure}[t]
\centering
\subfloat[]{\includegraphics[width=0.5\columnwidth, height=2cm]{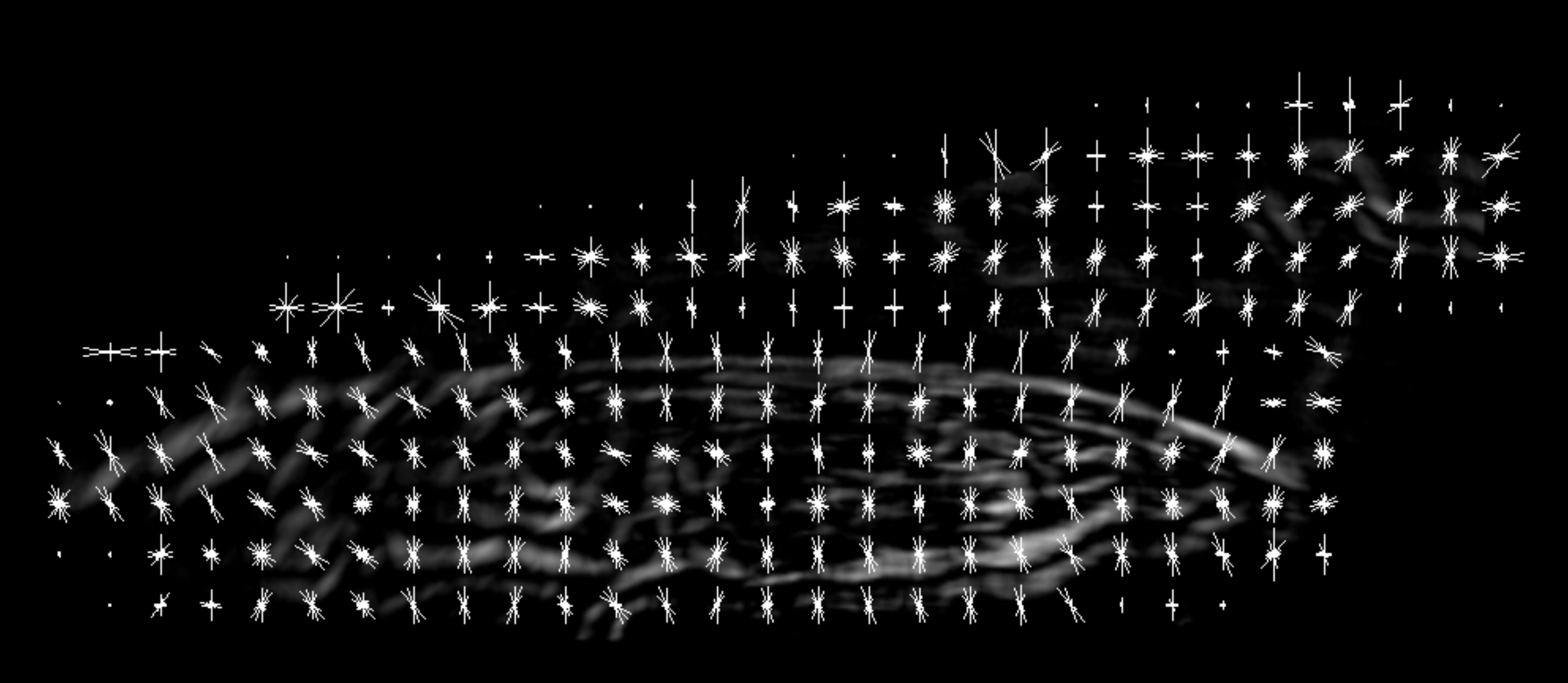}
\label{fig_hog_descriptor}}
\subfloat[]{\includegraphics[width=0.5\columnwidth, height=2cm]{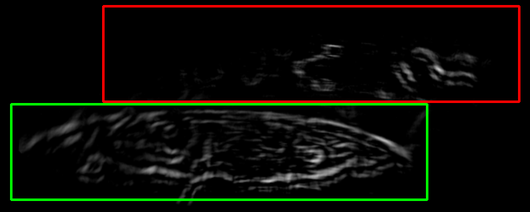}
\label{fig_negative_positive}}
\caption{\protect\subref{fig_hog_descriptor} HOG feature extraction. \protect\subref{fig_negative_positive} Positive (green rectangle) and negative (red rectangle) samples in the training data set.}
\end{figure}

Since the assets inspected by the AUV are all previously known, the target detection algorithm is trained via supervised learning. HOG descriptor is not rotation invariant, so that the target orientation is manually informed by selecting the heading of the target during the annotation process.

The training data was divided into positive and negative sets, wherein the former contains the target object. Before the training stage, the target orientation is normalized according to the target head annotated during the annotation step. To achieve that, all image from the training set is rotated to make the target head point be in 180 degrees with respect to the image vertical axis.

Two sets of vectors are created, one with the features extracted from the positive images, and another for negative images. These sets feed a linear SVM, in the training stage. A positive and negative example are shown in Fig. \ref{fig_negative_positive}, where the positive is in the green rectangle, and the negative is in the red rectangle. To extract the negative examples, we scan the image using a window with the same size of the annotated bounding box. The windows that do not overlap the annotated area are selected as negative example. If the image does not have a target object annotated, a window with size of $208\times48$ are used.

\begin{figure*}[t]
\centering
\subfloat[8$\times$8 step]{\includegraphics[width=0.38\textwidth]{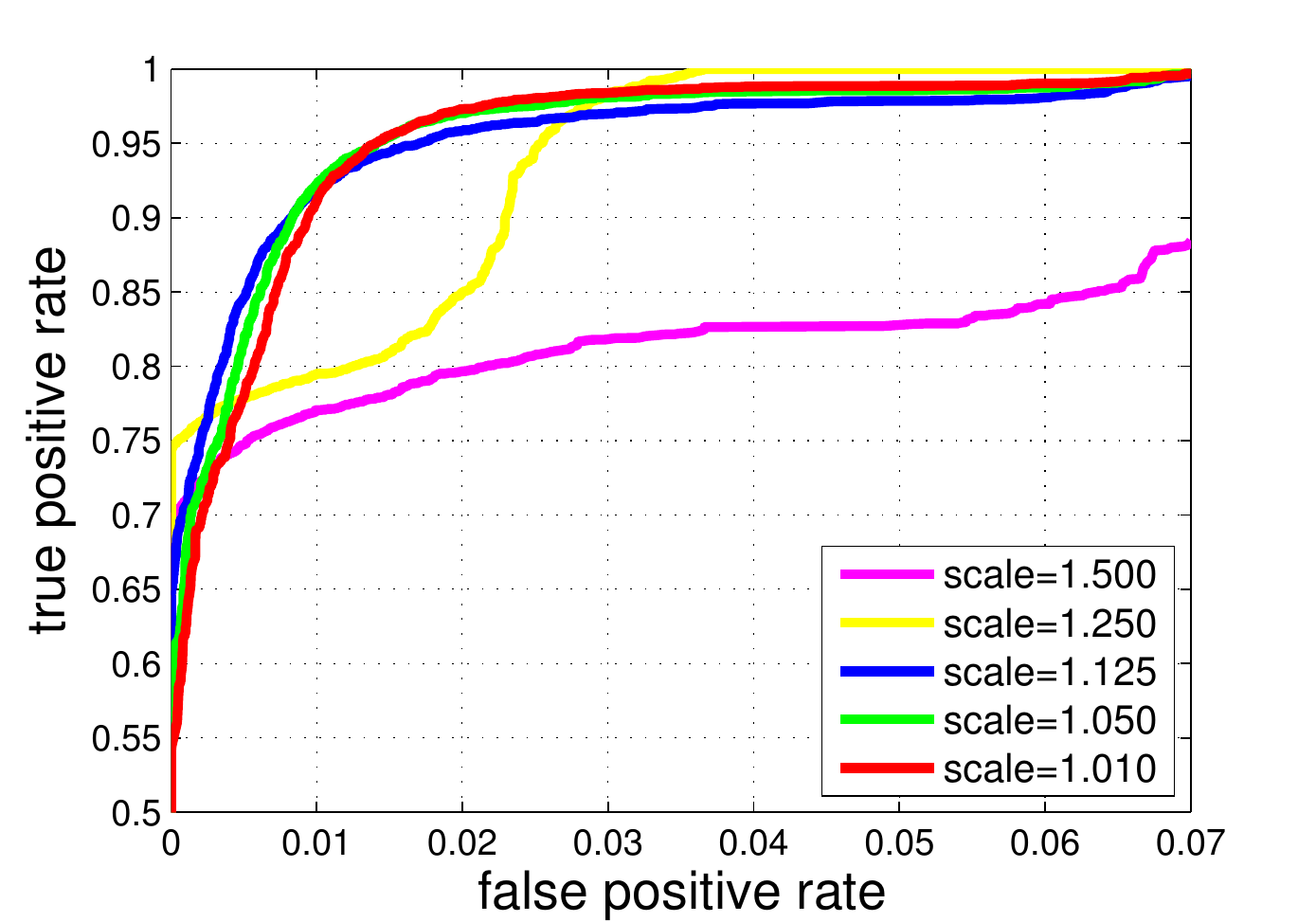}
\label{fig_roc_step_8x8}}
\subfloat[16$\times$16 step]{\includegraphics[width=0.38\textwidth]{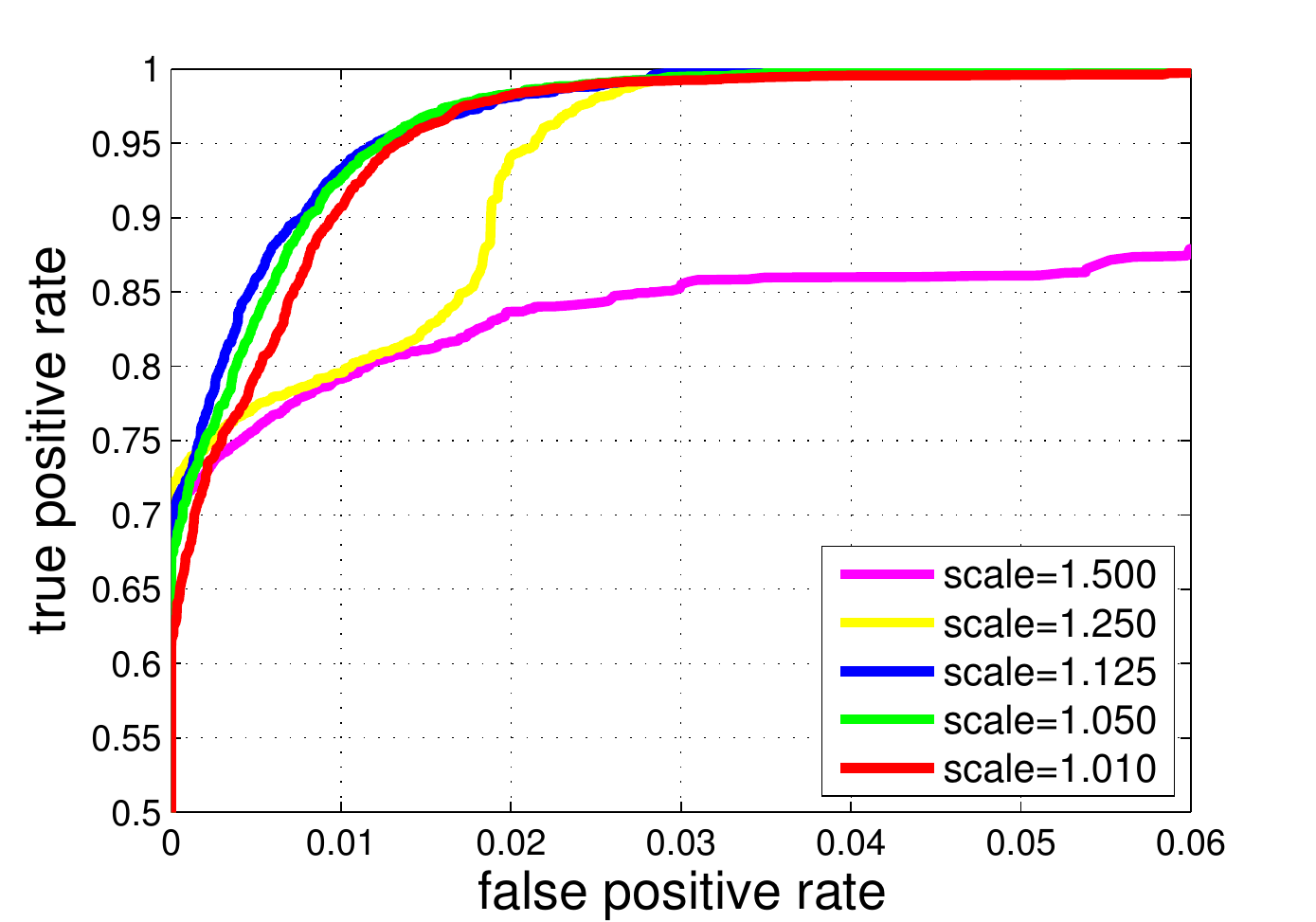}
\label{fig_roc_step_16x16}}
\caption{ROC curves for \textit{"Vapor da Jequitaia"} detector.
\protect\subref{fig_roc_step_8x8} sliding window step of 8x8 pixels,
\protect\subref{fig_roc_step_16x16} sliding window step of 16x16 pixels
}
\label{fig_curves}
\end{figure*}

\subsection{Searching and recognizing the target object}
\label{sec_searching_target}

The used forward-looking sonar has an adjustable range, consequently making the target object to have different sizes in the sonar image. As the HOG descriptor is not scale-invariant, the image pyramid method is used to find objects at different scales. A multi-scale orientation detector searches for the target by means of a sliding window combined with an image pyramid-based search method, over different image orientations. This method moves a fixed size rectangle over the image from top-left to bottom-right of the sonar image. For each window, a linear SVM is applied to determine if the window contains the target object or not.

To guarantee the rotation invariance, the sonar image must be rotated into different orientations. For each orientation, the multi-scale detector is executed and results are saved into a list. Each item in the list contains the SVM weight, the window that may contain the target object and the image orientation. The item of the list with the largest SVM weight is selected. The corresponding image orientation is used to rotate the window into the default orientation, resulting in a standardized view of the target object.

\section{Experimental Results}

\subsection{Data acquisition}
\label{sec_data_acquisition}

Data acquisition was carried out using the FlatFish robot \cite{Albiez2015}. FlatFish is a sub-sea resident AUV, which was built to perform on-demand close visual structural inspection at oil and gas sites. The robot is equipped with a Tritech Gemini 720i sonar (under the robot), as an acoustic global navigation sensor, which operates at 720$kHz$, with $120^{\circ}$ horizontal and $20^{\circ}$ beamwidths, and a downward of $-10^{\circ}$. Across the horizontal axis, the system is comprised of 256 beams with effective azimuth-angular beam resolution of $0.5^{\circ}$. The coverage range varies between $0.2-120m$, with a frame rate up to $30Hz$. During the acquisition process, the sonar range was set to 30 and 35 meters, and the AUV was moved surrounding the shipwreck to simulate the inspection task. The FlatFish was manually controlled for the sway and surge degrees, being autonomous for yaw and heave degree.

The subsea experiments were conducted at \textit{Todos os Santos Bay, Salvador, Bahia}. The main target was \textit{Vapor da Jequitaia} -- a vessel shipwrecked in 1905, which lies approximately 7 meters deep in the water. The \textit{Vapor da Jequitaia} has 27-meter long and has a distinctive shape, and even with a great number of holes in its hull, it can be inspected from the top, turning to be a very suitable testing target (see Fig. \ref{fig_flat_fish_and_shipwreck}).

\subsection{Data preparation: training and test sets}
\label{sec_training_data}

The training data was prepared using the steps described in Section \ref{sec_learning_process}. 431 acoustic images were gathered and annotated to form the training data set. Due to the difficulty to find a good target in the sea environment, we used just \textit{Vapor da Jequitaia} in our experiments, that represents our target object. These images were resized to fit the annotated bounding box into the detection window, whose size is $208\times48$ pixels, keeping the \textit{Vapor da Jequitaia} aspect ratio. We extracted 431 positive examples, and 469 negative examples from the training dataset. For our experiments, we used a linear SVM, with $C=0.01$. 

A dataset with 1222 acoustic images containing \textit{Vapor da Jequitaia} was annotated to assess the performance of our method. For each result given by the linear SVM classifier, we compare the detected area to the ground truth. The ROC curves in the Figs. \ref{fig_roc_step_8x8} and \ref{fig_roc_step_16x16} show the detector performance, which is quantified by a \textbf{true positive rate} (defined as $\frac{tp}{tp+fp}$, where $tp$ and $fp$ denotes true positive and false positive areas), and the \textbf{false positive rate} (defined as $\frac{fp}{fp+tn}$, where $tn$ denotes the true negative area).

One of the parameters of our multi-scale orientation detector is the sliding window step size, which defines how many pixels will be skipped during the sliding window. As shown in Figs. \ref{fig_roc_step_8x8} and \ref{fig_roc_step_16x16}, we tested the sliding window step size with values of $8\times8$ and $16\times16$ pixels. Scale indicates how much of the image will be resized in the image pyramid representation. Five different scales were evaluated (see Figs. \ref{fig_roc_step_8x8} and \ref{fig_roc_step_16x16}). Testing images were rotated from 0 to 180$^{o}$ in steps of 10$^{o}$ to search the target in different orientations. Windows with the highest SVM weight was chosen. 

\begin{table}[t]
\centering
\renewcommand{\arraystretch}{1.3}
\caption{Vapor da Jequitaia Detection Evaluation. TPR and FPR denote true positive rate and false positive rate, respectively.}
\label{tab_detector_accuracy}
\scriptsize
\begin{tabular}{|c|c|c|c|c|c|c|}
\hline
\multirow{2}{*}[-5pt]{Scale} & \multicolumn{2}{c|}{Step 8x8} & \multicolumn{2}{c|}{Step 16x16} \\ \cline{2-5} &
\begin{tabular}[c]{@{}c@{}}TPR \\ Average (\%)\end{tabular} &
\begin{tabular}[c]{@{}c@{}}FPR \\ Average  (\%)\end{tabular} &
\begin{tabular}[c]{@{}c@{}}TPR \\ Average (\%)\end{tabular} &
\begin{tabular}[c]{@{}c@{}}FPR \\ Average (\%)\end{tabular} \\
\hline
1.010 & 87.3 & 1.7 & 87.6 & 1.7 \\
\hline
1.050 & 85.3 & 1.4 & 83.3 & 1.2 \\
\hline
1.125 & 82.8 & 1.2 & 77.6 & 0.8 \\
\hline
1.250 & 76.9 & 1.5 & 71.2 & 1.2 \\
\hline
1.500 & 79.2 & 3.8 & 63.8 & 0.9 \\
\hline
\end{tabular}
\end{table}

Table \ref{tab_detector_accuracy} summarizes accuracy evaluation in our experiments. \textbf{TPR average} and \textbf{FPR average} denote the averages of the true positives and false positives rates calculated in the test stage, respectively. As shown in Table \ref{tab_detector_accuracy}, scale $1.01$ presents the best result in both window step size with the best TPR average of $87.6$\%, using the window step of $16\times16$ pixels. For all scales, low values of false positive rate were computed with the highest obtained value equal to $1.7\%$. As the scale value increases, the TPR average decreases, and thus the detector performance is reduced.

\section{Conclusion and outlook}

A novel trainable method to detect and recognize specific submerged objects with a forward-looking sonar was presented here. By taking advantages of specific image processing techniques, we reduced the sensor noise and the seabed background of the sonar image, emphasizing the shape and the borders of the target inspection object. Using multi-scaled orientation detector and a linear SVM, the target object is detected and recognized in a rotation-invariant way. The capacity of our method to detect the known objects has been measured by means of the real data collected by Tritech Gemini 720i sonar. The very main goal was to support the Flatfish with under-the-sea inspection. Ongoing work have been carrying on to fully integrate the proposed method with the FlatFish's navigation system. The goal is to allow controlling the vehicle with respect to the target inspection object, trying to make the inspection task more accurate. Future work includes increasing the number of target objects, integrating the recognition system with a tracking method.



\end{document}